# Susceptibility of Texture Measures to Noise: An Application to Lung Tumor CT Images

Omar S. Al-Kadi*, *Member, IEEE* and D. Watson

*Abstract*—Five different texture methods are used to investigate their susceptibility to subtle noise occurring in lung tumor Computed Tomography (CT) images caused by acquisition and reconstruction deficiencies. Noise of Gaussian and Rayleigh distributions with varying mean and variance was encountered in the analyzed CT images. Fisher and Bhattacharyya distance measures were used to differentiate between an original extracted lung tumor region of interest (ROI) with a filtered and noisy reconstructed versions. Through examining the texture characteristics of the lung tumor areas by five different texture measures, it was determined that the autocovariance measure was least affected and the gray level co-occurrence matrix was the most affected by noise. Depending on the selected ROI size, it was concluded that the number of extracted features from each texture measure increases susceptibility to noise.

## I. INTRODUCTION

Texture in medical images can offer an important source of information on the state of the health of an examined organ. Often medical images are degraded by different types and levels of noise, which might arise from photon, electronics and/or quantisation [1], affecting the fine structure of the examined texture in these images [2]. Therefore, having clear and relatively noise-free acquired images plays a significant role in medical image analysis.

Physicians tend to use computed texture measures from regions of interest (ROIs) for diagnosis purposes and for eventually choosing the appropriate treatment procedure. It has been shown that fractal analysis of lung tumors texture in Computed Tomography (CT) images can assist in distinguishing between aggressive and non-aggressive tumors [3]. However, we need to take into consideration when examining the texture of a small ROI in a medical image, that noise could adversely affect the accuracy of the measured texture parameters and cause errors in the reported diagnosis [4]. Many studies concerned with noise reduction and CT image enhancement have been taken [5, 6], yet this paper aims to provide a comparison study between five different well-known texture measures to investigate their susceptibility to noise occurring in CT images, which will give an indication of texture measure reliability and fidelity in analyzing medical images, with a possible expansion to other modalities.

## II. METHODS

First the type of noise needs to be identified, and then two images are generated from each original CT image, one with a reduced noise and another with an enhanced noise. These versions are CT reconstructed and two new ROIs ─ one from each of the two reconstructed versions ─ are extracted manually from the tumor area (see arrow 1 in Fig. 1) and compared with the original ROI according to five different texture measures. The process is summarized in Fig.2, and the used procedure is described in detail as follows:

### A. Image Acquisition

We used 11 DICOM (Digital Imaging and Communication in Medicine) CT images of lung tumors from 11 different patients (6 males and 5 females with age $63 \pm 8$ year old with lung cancer greater than 10 mm$^2$), having a resolution of 12 bits per pixel. The images were acquired with X-ray tube voltage and current of 140 kV and 200mAs, a 10 mm slice thickness with matrix size 512 x 512 and B reconstruction filter. All acquired images were ethically approved, and our work did not influence the diagnostic process or the patient's treatment.

### B. Noise Evaluation

The original image is first inspected for presence of noise, and the type of noise is appropriately identified for removal without destroying the fine structure of the image texture. Two new images will be produced from this phase, a clean (i.e. filtered original image) and distorted (i.e. the detected noise in the original image is doubled) versions.

*1) Noise Estimation*

A reasonably constant grey level area in the CT image is selected and checked for uniformity. The transverse section of the scanning table in the CT image was chosen for analysis (see arrow 2 in Fig. 1), and the histogram was plotted for it. Then the mean ($\mu$) and variance ($\sigma^2$) which were estimated from the plotted histogram are used to determine the parameters of three other types of noise probability density functions (PDFs) for their histograms to be plotted as well (see Table I). The selected noise types for this work were Gaussian, Rayleigh and Erlang [7]. Then the estimated histogram from the CT image will be matched against the generated noise PDFs to see to which one it best

This work was supported by the University of Jordan, Amman, Jordan, under Scholarship 8/2214/145. *Asterisk indicates corresponding author.*

*Omar S. Al-Kadi is with the Department of Informatics, University of Sussex, Brighton BN1 9QH, U.K. (phone: +44 1273 877971; fax: +44 1273 678195; e-mail: o.al-kadi@sussex.ac.uk).

D. Watson is with the Department of Informatics, University of Sussex, Brighton BN1 9QH, U.K. (e-mail: desw@sussex.ac.uk).

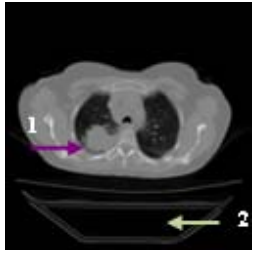

Fig. 1. Arrow 1: lung tumour area, arrow 2: transverse section in the scanning table used for noise estimation.

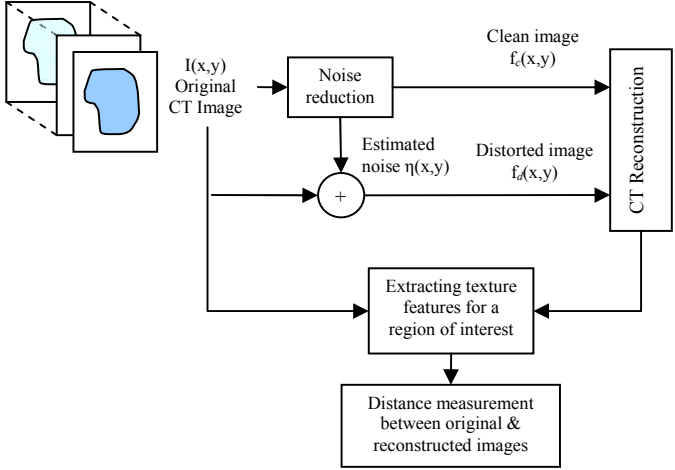

Fig. 2. Methodology used to assess texture measures' susceptibility to noise for lung tumor CT images.

corresponds. This process is carried out for all 11 images.

The intensities histograms obtained from the uniform areas had a shape resembling additive Gaussian and multiplicative Rayleigh noise PDFs with $\mu$ and $\sigma^2$ varying between 13.2 to 17.4 and 24.7 to 65.9; respectively. Matusita distance was used to compare between the original noise ($P_O$) and the three generated noise ($P_N$) distributions to see to which the measured noise is least deviated as shown in (1).

$$M(P_O, P_N) = \sqrt{\sum_i \left(\sqrt{P_O} - \sqrt{P_N}\right)^2} \quad (1)$$

Fig. 3 shows a histogram of noise obtained from one of the CT images compared to three different types of generated noise (Gaussian, Rayleigh and Erlang) using the estimated $\mu$ and $\sigma^2$. We can see for this case that the shape of the Rayleigh noise appears to resemble the CT noise histogram, and the distance measure supports this conclusion (see case 3 in Table II). Also in Table II, six of the examined cases showed a Rayleigh noise distribution while the rest appeared to have a Gaussian distribution. It was shown that if the standard deviation of the estimated noise is far less than the mean intensity, the noise will approach a Gaussian distribution, whilst if it is for greater than the mean intensity will give a Rayleigh distribution [8]. Nevertheless, the reason not having a single noise type in the analysed CT images even though the same CT scanner was used needs to be further investigated.

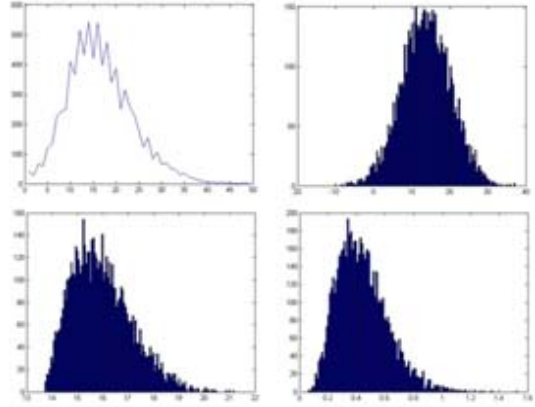

Fig. 3. From left to right and from top to bottom, histograms with $\mu_\eta = 13.6977$, $\sigma_\eta^2 = 41.1472$ of transverse section of scanning table in CT images followed by corresponding generated Gaussian, Rayleigh and Erlang noises; respectively.

*2) Adaptive Filtering*

Having identified the type of noise, we need to clean each of the CT images given the corresponding noise variance. As the tumour area is relatively small as compared to the total image size, we need an adaptive filter that reduces local noise and preserves the edges and fine structures in the CT image for subsequent accurate analysis. An adaptive filter ($S_{xy}$) of size 5 x 5 which covers nearly 1% of the image in each step is used for local noise reduction. Its behaviour changes adaptively depending on the statistical characteristics of the region inside the filter as defined in the following formula [7]:

$$f_c(x, y) = I(x, y) - \frac{\sigma_\eta^2}{\sigma_L^2}\left[I(x, y) - \mu_L\right] \quad (2)$$

Here $I(x,y)$ is the value of the original image suspected to have subtle noise at $(x,y)$; $\sigma_\eta^2$ the variance of the noise corrupting $f_c(x,y)$ to form $I(x,y)$; $\mu_L$ is the local mean of the pixels in $S_{xy}$; and $\sigma_L^2$, the local variance of the pixels in $S_{xy}$. In case of noise absence (i.e. $\sigma_\eta^2 = 0$) the filter will return the original image. Also it preserves the edges in case the local variance is high. If noise and local variances are equal the filter returns the arithmetic mean value of the pixels in $S_{xy}$.

In order to study the impact of increased noise on texture analysis measures used in CT images, a distorted image $f_d(x,y)$ is generated by simply adding the estimated noise $\eta(x,y)$ — which is a by-product of the adaptive filtering process — to the original image $I(x,y)$, as in (3).

$$f_d(x, y) = I(x, y) + \eta(x, y) \quad (3)$$

TABLE I
PDFS FOR THREE DIFFERENT TYPES OF NOISE AND THEIR CORRESPONDING MEAN AND VARIANCE

| Noise | Gaussian | Rayleigh | Erlang |
|---|---|---|---|
| PDF | $P_z(z) = \frac{1}{\sqrt{2\pi}b}\exp(-\frac{(z-a)^2}{2b^2})$ where $-\infty < z < \infty$ | $P_z(z) = \begin{cases} \frac{2}{b}(z-a)\exp(-\frac{(z-a)^2}{b}) & z \geq a \\ 0 & z < a \end{cases}$ | $P_z(z) = \frac{a^b z^{b-1}}{(b-1)!}\exp(-az)$ where $z \geq 0$ |
| Mean & Variance | $\mu = a$, $\sigma^2 = b^2$ | $\mu = a$, $\sigma^2 = b^2$ | $\mu = a$, $\sigma^2 = \frac{b}{a^2}$ |

TABLE II
MATUSITA DISTANCE BETWEEN ORIGINAL EXTRACTED UNIFORM LUNG TUMOR ROIS AND THREE TYPES OF NOISE DISTRIBUTIONS.

| CT Image ROI | Generated noise | | |
|---|---|---|---|
| | Gaussian | Rayleigh | Erlang |
| Case1 | 0.3091 | 0.5578 | 0.6001 |
| Case2 | 0.1611 | 0.5681 | 0.7889 |
| Case3 | 0.5181 | 0.2855 | 0.6115 |
| Case4 | 0.1646 | 0.4927 | 0.9515 |
| Case5 | 0.3359 | 0.5238 | 0.4315 |
| Case6 | 0.3616 | 0.6888 | 0.5170 |
| Case7 | 0.6601 | 0.1967 | 0.5712 |
| Case8 | 0.4542 | 0.3016 | 0.7447 |
| Case9 | 0.6217 | 0.2311 | 0.6211 |
| Case10 | 0.4069 | 0.3255 | 0.7019 |
| Case11 | 0.3971 | 0.3219 | 0.6046 |

## III. CT IMAGE RECONSTRUCTION

An open-source software called CTSim [9] was used in the simulation process to reconstruct the CT images. The software simulates the process of collecting X-ray data of phantom objects. We considered the intensity of each pixel in the original DICOM CT image as a rectangle object of unit distance representing the X-ray attenuation coefficient referring to that position. By the end of this stage, we have three different CT images for each case, which are the original and two versions acquired under different conditions. Texture analysis is then performed on the 33 CT images as described in the next section.

## IV. TEXTURE FEATURE EXTRACTION

As different lung tumors vary in size depending on the stage of development and aggression, a size that ensures capturing of the texture variation in each ROI is needed. Smaller areas would not have sufficient pixels to reliably compute the texture parameters, while larger areas would exclude relatively small size tumors from calculations. Therefore, we have empirically chosen an ROI of size 32 x 32 pixels to be extracted from each tumor regions of the 33 CT images as this chosen size would balance the trade-off between tumor size and texture area. Five different texture analyses methods used in [10] were applied to analyze the texture characteristics of the ROIs. These methods are represented by Gaussian Markov random field (GMRF) and fractal dimension (FD) which are model based, and autocovariance function (ACF), runlength matrix (RLM) and grey level co-occurrence matrix (GLCM) which are statistical based.

## V. DISTANCE MEASUREMENTS

The final phase in this work is the comparison process where the reconstructed images are compared to the original CT image in terms of how much deviation is incurred in the reconstructed images due to noise (removal/addition) after normalising all extracted texture measures. Two non-parametric statistical distance measures were used for comparison. Although these distance measures are often used in determining accuracy of clusters separability, they are used here to indicate how non-separable (i.e. close) the reconstructed images are to the original. Our aim is to find the best non-separable texture measure between the original and reconstructed images which is less susceptible to noise.

### A. Fisher criterion

The Fisher criterion is a nonparametric measure used to assess the quality of separability of two classes. It represents the ratio of the between-class variance relative to the within-class variance. In case of a multi-feature vector, the distance can be measured by the formula [11]:

$$J(w) = \frac{W^T S_B W}{W^T S_w W} \qquad (4)$$

Where $S_B$ and $S_W$ are the between-class and within-class scatter matrices. For our case smaller values show better performance since the larger the Fisher criterion values the more significant the difference between the two assessed classes.

### B. Bhattacharyya Error Bound

This method calculates the upper bound of classification error between feature class pairs [11]. In our case larger error values are better since it shows that both the original and reconstructed images are less separable (i.e. nearly identical).

$$B_{c_1 c_2} = \frac{1}{8}(\mu_1 - \mu_2)^T \left(\frac{\Sigma_1 + \Sigma_2}{2}\right)^{-1} (\mu_1 - \mu_2) + \frac{1}{2}\ln\left(\frac{|\Sigma_1 + \Sigma_2|}{2\sqrt{|\Sigma_1||\Sigma_2|}}\right) \qquad (5)$$

where $|\Sigma_i|$ is the determinant of $\Sigma_i$, and $\mu_i$ and $\Sigma_i$ are the mean vector and covariance matrix of class $C_i$.

TABLE III
FISHER AND BHATTACHARYYA DISTANCE BETWEEN ORIGINAL AND
RECONSTRUCTED CLEAN AND NOISY CT IMAGES

| Method | features | $F_{oc}$ | $F_{on}$ | $B_{oc}$ | $B_{on}$ |
|--------|----------|----------|----------|----------|----------|
| ACF    | 8        | 4.03E-30 | 2.45E-30 | -44.59   | -44.74   |
| FD     | 5        | 2.67E-29 | 7.34E-28 | -52.90   | -57.75   |
| RLM    | 16       | 4.63E-26 | 8.82E-26 | -201.89  | -218.23  |
| GMRF   | 13       | 6.71E-24 | 1.13E-23 | -170.36  | -175.10  |
| GLCM   | 32       | 3.35E-12 | 4.10E-14 | -Inf     | -Inf     |

## VI. RESULTS AND DISCUSSION

In Table III, $F_{oc}$ and $F_{on}$ are the differences between the original CT image and the reconstructed clean and distorted images measured by Fisher distance, respectively. Similarly for $B_{oc}$ and $B_{on}$ in the last and next to last columns but this time measured by Bhattacharyya distance. For the Fisher criterion the ACF was the least affected by noise, followed by FD then RLM, GMRF and GLCM, respectively. Similar results were obtained using the Bhattacharyya distance test, but the GMRF was less affected by noise as compared to RLM. Also, the Bhattacharyya distance showed that the clean CT reconstructions are much nearer to the original from the dirty ones, therefore adaptive filtering can assist in improving accuracy.

It seems that the number of extracted features by each texture measures plays an important role in susceptibility to noise. GLCM which extracts 32 different features was more prone to noise as compared to the ACF which has 8 features. This might be due to the fact that texture measures with large number of features tend to capture more variations of the intensity, and as a result the probability of noise contribution would be amplified. On the other hand, although some studies reported Gaussian noise distributions in low dose CT mages [12], this paper showed that other type of noise than Gaussian can be encountered even when using the same CT scanner.

This indicates that noise can have some impact on the variability of diagnosis reports depending on the used texture measure for analysis and classification. Some texture measure are more reliable in terms of classification [13], yet their accuracy might start to give misleading results in case of noise presence, causing an increase in inaccuracy as noise becomes more obvious. Therefore, accuracy and noise susceptibility must be taken into consideration by the physician depending on the type of analysis and the area of texture. Taking into consideration the acquired image resolution, physicians can use texture measures such as FD or ACF for small areas (e.g. size ≤ 32 x 32 pixels for image resolution used in this study) of texture where the probability of noise deforming the structure of the texture is higher, and use GLCM, GMRF or RLM for larger ROIs. Moreover, filtering noisy CT images with an adaptive filter can assist in better analysis and classification.

## VII. CONCLUSION

This work investigated the susceptibility of five different texture analysis measures to noise by using two distance measurement methods to compare the original CT images with their corresponding reconstructed clean and noisy versions. It was shown that the texture measures with few features such as the ACF and FD was the least affected by noise in both distance tests as compared to GLCM which had the highest number of features. Also adaptively filtered images can assist in reducing subtle noise, and hence offer better texture accuracy. The methodology used in this paper is being applied to a different set of lung tumor CT images acquired by a different CT manufacturer. Also the effects of other texture analysis methods such as Gabor filters and wavelets and for other modalities (MR and ultrasound) are being investigated.

ACKNOWLEDGMENT

We would like to thank the Clinical Imaging Science Centre at the University of Sussex for provision of the CT images used in this paper.